\documentclass[sigconf]{acmart}

\usepackage{subcaption}
\usepackage[capitalise, noabbrev]{cleveref}

\AtBeginDocument{%
  }

\copyrightyear{2024}
\acmYear{2024}
\setcopyright{rightsretained}
\acmConference[UbiComp Companion '24]{Companion of the 2024 ACM International Joint Conference on Pervasive and Ubiquitous Computing Pervasive and Ubiquitous Computing}{October 5--9, 2024}{Melbourne, VIC, Australia}
\acmBooktitle{Companion of the 2024 ACM International Joint Conference on Pervasive and Ubiquitous Computing Pervasive and Ubiquitous Computing (UbiComp Companion '24), October 5--9, 2024, Melbourne, VIC, Australia}
\acmDOI{XXXXXXX.XXXXXXX}

\acmISBN{}

\begin{document}

%%
%% The "title" command has an optional parameter,
%% allowing the author to define a "short title" to be used in page headers.
\title{Leveraging Hybrid Intelligence Towards Sustainable and Energy-Efficient Machine Learning}

%%
%% The "author" command and its associated commands are used to define
%% the authors and their affiliations.
%% Of note is the shared affiliation of the first two authors, and the
%% "authornote" and "authornotemark" commands
%% used to denote shared contribution to the research.
%% used to denote shared contribution to the research.
\author{Daniel Geißler}
\email{daniel.geissler@dfki.de}
\orcid{0000-0003-2643-4504}
\affiliation{%
  \institution{DFKI}
  \city{Kaiserslautern}
  \country{Germany}
}

\author{Paul Lukowicz}
\email{paul.lukowicz@dfki.de}
\orcid{0000-0003-0320-6656}
\affiliation{%
  \institution{DFKI}
  \country{}
}
\affiliation{%
  \institution{University of Kaiserslautern-Landau}
  \city{Kaiserslautern}
  \country{Germany}
}

%%
%% By default, the full list of authors will be used in the page
%% headers. Often, this list is too long, and will overlap
%% other information printed in the page headers. This command allows
%% the author to define a more concise list
%% of authors' names for this purpose.
\renewcommand{\shortauthors}{Geißler et al.}

%%
%% The abstract is a short summary of the work to be presented in the
%% article.
\begin{abstract}

Hybrid intelligence aims to enhance decision-making, problem-solving, and overall system performance by combining the strengths of both, human cognitive abilities and artificial intelligence.
With the rise of Large Language Models (LLM), progressively participating as smart agents to accelerate machine learning development, Hybrid Intelligence is becoming an increasingly important topic for effective interaction between humans and machines.
This paper presents an approach to leverage Hybrid Intelligence towards sustainable and energy-aware machine learning.
When developing machine learning models, final model performance commonly rules the optimization process while the efficiency of the process itself is often neglected.
Moreover, in recent times, energy efficiency has become equally crucial due to the significant environmental impact of complex and large-scale computational processes.
The contribution of this work covers the interactive inclusion of secondary knowledge sources through Human-in-the-loop (HITL) and LLM agents to stress out and further resolve inefficiencies in the machine learning development process.

\end{abstract}

%%
%% The code below is generated by the tool at http://dl.acm.org/ccs.cfm.
%% Please copy and paste the code instead of the example below.
%%

\begin{CCSXML}
<ccs2012>
   <concept>
       <concept_id>10003120.10003145.10003151</concept_id>
       <concept_desc>Human-centered computing~Visualization systems and tools</concept_desc>
       <concept_significance>500</concept_significance>
       </concept>
   <concept>
       <concept_id>10003120.10003121.10003129</concept_id>
       <concept_desc>Human-centered computing~Interactive systems and tools</concept_desc>
       <concept_significance>500</concept_significance>
       </concept>
   <concept>
       <concept_id>10003120.10003123.10011760</concept_id>
       <concept_desc>Human-centered computing~Systems and tools for interaction design</concept_desc>
       <concept_significance>500</concept_significance>
       </concept>
 </ccs2012>
\end{CCSXML}

\ccsdesc[500]{Human-centered computing~Visualization systems and tools}
\ccsdesc[500]{Human-centered computing~Interactive systems and tools}
\ccsdesc[500]{Human-centered computing~Systems and tools for interaction design}

%%
%% Keywords. The author(s) should pick words that accurately describe
%% the work being presented. Separate the keywords with commas.
\keywords{Hybrid Intelligence, Interactivity, Sustainability, Energy Awareness}
%% A "teaser" image appears between the author and affiliation
%% information and the body of the document, and typically spans the
%% page.
% \begin{teaserfigure}
%   \includegraphics[width=\textwidth]{sampleteaser}
%   \caption{Seattle Mariners at Spring Training, 2010.}
%   \Description{Enjoying the baseball game from the third-base
%   seats. Ichiro Suzuki preparing to bat.}
%   \label{fig:teaser}
% \end{teaserfigure}

%%
%% This command processes the author and affiliation and title
%% information and builds the first part of the formatted document.
\maketitle
\section{Problem Statement}

The field of sustainable machine learning has witnessed increasing relevance in recent years, driven by the fast-paced expansion of machine learning applications in pervasive systems \cite{yao2023machine}.
Even though the efficiency, considering hardware from low-power embedded devices to high-performance clusters, has improved over time, it cannot keep up with the rapidly increasing amount of devices that are utilized in our daily lives \cite{desislavov2023trends}.
When it comes to energy-efficient optimization, research commonly focuses on optimizing for time reduction instead of energy \cite{mehlin2023towards}.
However, shrinking time commonly requires even more computational resources to counteract.

From the data-level perspective, the quality of training data significantly impacts the final performance of machine learning models \cite{Soni2023Evaluating}. 
Small nuances in data quality, such as issues arising from sensor quality, placement, and annotation, can drastically affect the overall model performance, especially in the fields of Human Activity Recognition HAR \cite{zhang2022deep}. 
These imperfections lead to inefficient utilization of resources, as weak data quality necessitates additional training cycles and resource expenditure to achieve acceptable performance levels.
Moreover, common approaches to address these inefficiencies, like enlarging the model architectural complexity to improve generalization, often result in an even greater waste of resources due to unawareness of the causes \cite{li2020train}. 

Another significant issue is the lack of a clear framework for measuring energy consumption in machine learning training \cite{10171575}. 
Many existing approaches only track parts of the system, such as GPU usage, without considering the overall energy footprint of the entire infrastructure \cite{henderson2020towards}. 
Work in this direction needs to cover the custom hardware landscape to offer more precise measurements and obtain guidelines on how to efficiently utilize resources.

To improve the sustainability of machine learning practices, it is essential to not only determine what aspects of the training process are inefficient but also to understand why they are inefficient \cite{alicioglu2022survey}. 
This requires a holistic examination of the hardware, data quality, and model architecture in conjunction to cover the potential abstraction levels for optimization.
Hybrid Intelligence offers means for the integration of additional knowledge through visual and numeric analysis to incorporate computational resource savings within the model optimization process \cite{akata2020research}.

\section{Related Work}
Several studies focus on the either enhancing energy awareness or improving the interaction and visualization of machine learning development as presented in the following sections.
To the best of our knowledge, there is currently no work combining these two areas into one coherent topic.
%No works to incorporate energy into the training and or reduce consumption through smart guiding of humans or agents

\subsection{Energy Tracking}

In the AI sector, energy consumption tools are still a niche, especially when it comes to generating awareness about the energy expenditure for training machine learning models \cite{mobius2013power}. 
They are usually designed to capture energy information by building an additional layer between the system’s hardware configuration and the user’s model training process \cite{henderson2020towards}. 
The power consumption of the GPU is considered the largest part of the training process as it performs the core work with the parallel processing of mathematical tasks \cite{mittal2014survey}. 
Nevertheless, other hardware components and even secondary power consumers like the cooling system or the power supply unit itself contribute to the overall energy consumption.

Works like Carbontracker \cite{anthony2020carbontracker}, eco2Ai \cite{budennyy2022eco2ai} and Green Algorithms \cite{lannelongue2021green} utilize this approach to gather data from the hardware.
To handle missing elements in the calculations, software like Carbontracker multiplies its results with an efficiency constant to incorporate untracked secondary power needs and efficiency losses.
With an extended focus on user experience, projects like Cloud Carbon \cite{cloudCarbon} or CodeCarbon \cite{CodeCarbon} extend the gathered knowledge and present it in analytic-based dashboards. 
Based on the calculated energy consumption and the user’s location, the average local energy mix from fossil and renewable energy sources is utilized to quantify the carbon emissions \cite{lacoste2019quantifying}.

\subsection{Interactive Machine Learning Visualizations}

Interactive visualizations can open up the black-box nature of machine learning models, making them more interpretable by allowing users to interrogate, explain, and even validate the process \cite{Hurley2021Interactive}.
Various tools are available, essentially aiming to enhance trust in machine learning models through increasing human perception of the model's internals \cite{Chatzimparmpas2020The}.
TensorFlow Graph Visualizer for instance helps users understand complex machine learning architectures by visualizing their underlying dataflow graphs, making it easier to debug and improve model structures \cite{Wongsuphasawat2018Visualizing}.
However, it is still a challenge to select the right visualization, especially when lossy dimension reduction strategies are applied and thus information content is missing \cite{Rauber2017Visualizing}.

Further, the interactivity of visualizations can elevate the insights and feed the user's input back into the model for adaptation \cite{Fuchs2009Visual}.
Works like \cite{Sacha2017What} have proven the advantages of Human-centered and HITL systems to enhance machine learning development through human knowledge integration and feedback.
Similar to the idea of this work, \cite{wei2022fine} proposes a human-in-the-loop approach to enhance deep neural network classification accuracy by leveraging human knowledge. 
By projecting high-dimensional latent spaces onto a two-dimensional workspace, users can interactively modify coordinates, which are then fed back to fine-tune the network, improving classification results.

\section{Methodology}

\begin{figure}[t]
  \centering
    \includegraphics[width=0.95\linewidth]{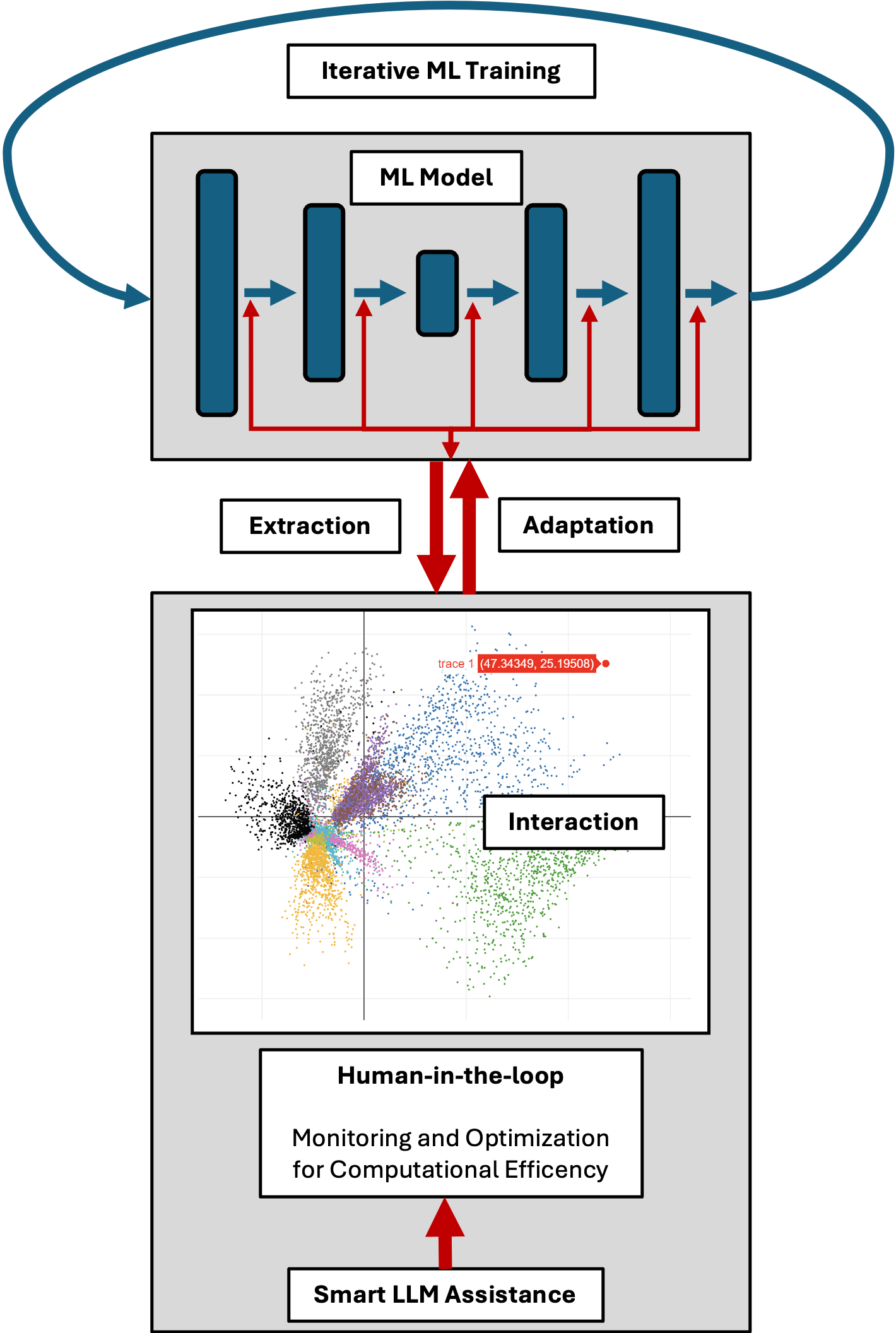}
    \caption{The concept of embedding Hybrid Intelligence into the traditional Machine Learning training to enhance energy efficiency from a data-driven perspective. }
    \vspace{-10pt}
    \label{fig:concept}
\end{figure}

The proposed concept leverages hybrid intelligence to elevate sustainable machine learning development. 
By incorporating HITL systems and LLMs as smart agent assistance, we aim to enhance resource efficiency and significantly reduce energy consumption. 
This approach ensures that machine learning models achieve performance goals without compromising on environmental considerations. 
The integration of Hybrid Intelligence support into machine learning workflows represents a promising solution to existing challenges, leading to more effective and environmentally responsible practices.
The following sections introduce the ongoing research steps to connect the relevant areas into a coherent framework.

\subsection{Hybrid Intelligence}

\begin{figure}[t]
  \centering
  \begin{subfigure}[b]{\linewidth}
    \centering
    \includegraphics[width=0.9\textwidth]{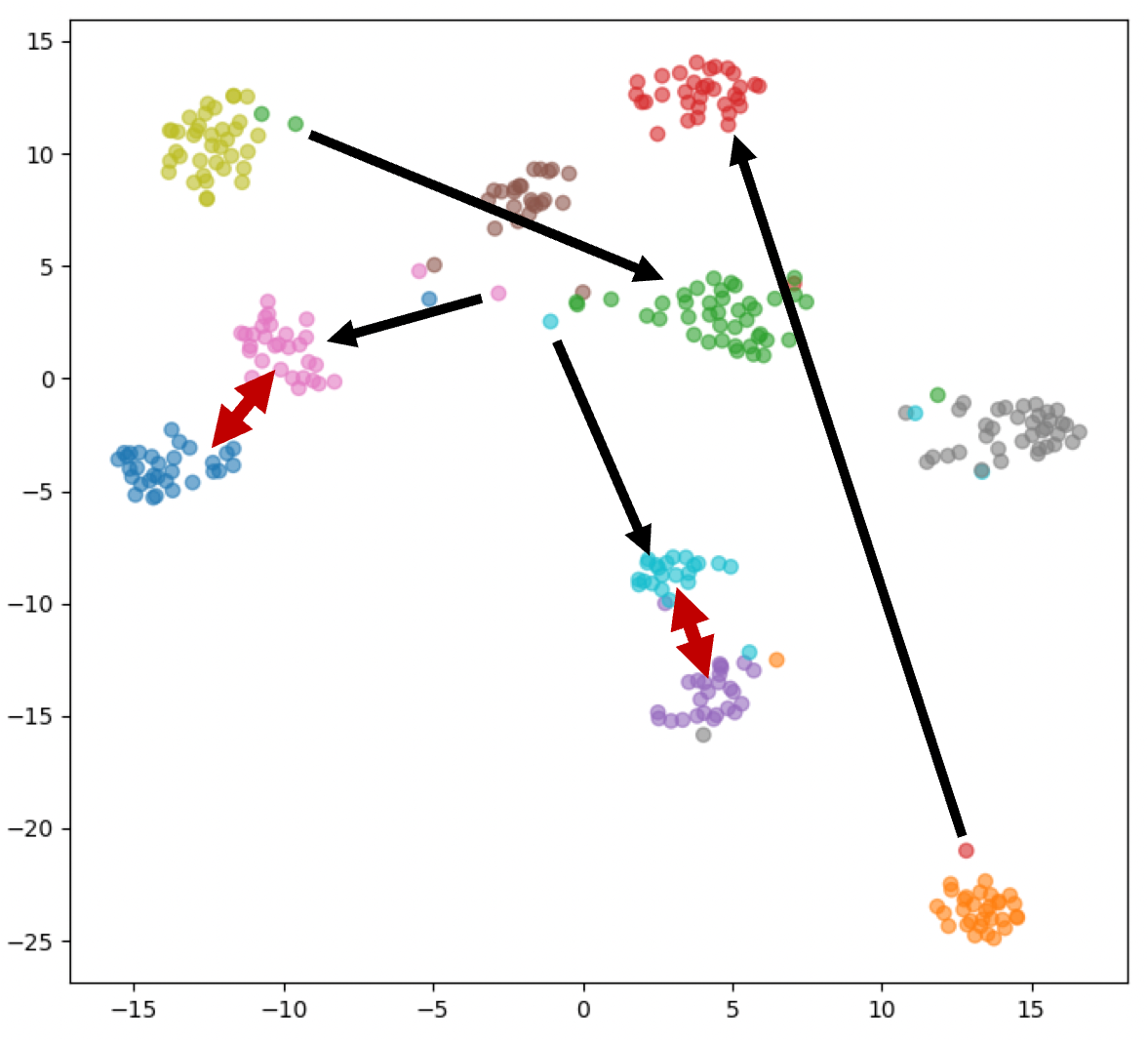}
    \caption{Scatter plot to rejoin outliers with their desired clusters (black arrow) or increase the distance between whole clusters (red arrow).}
    \vspace{10pt}
    \label{fig:scatter}
  \end{subfigure}
  \hfill
  \begin{subfigure}[b]{\linewidth}
    \centering
    \includegraphics[width=0.9\textwidth]{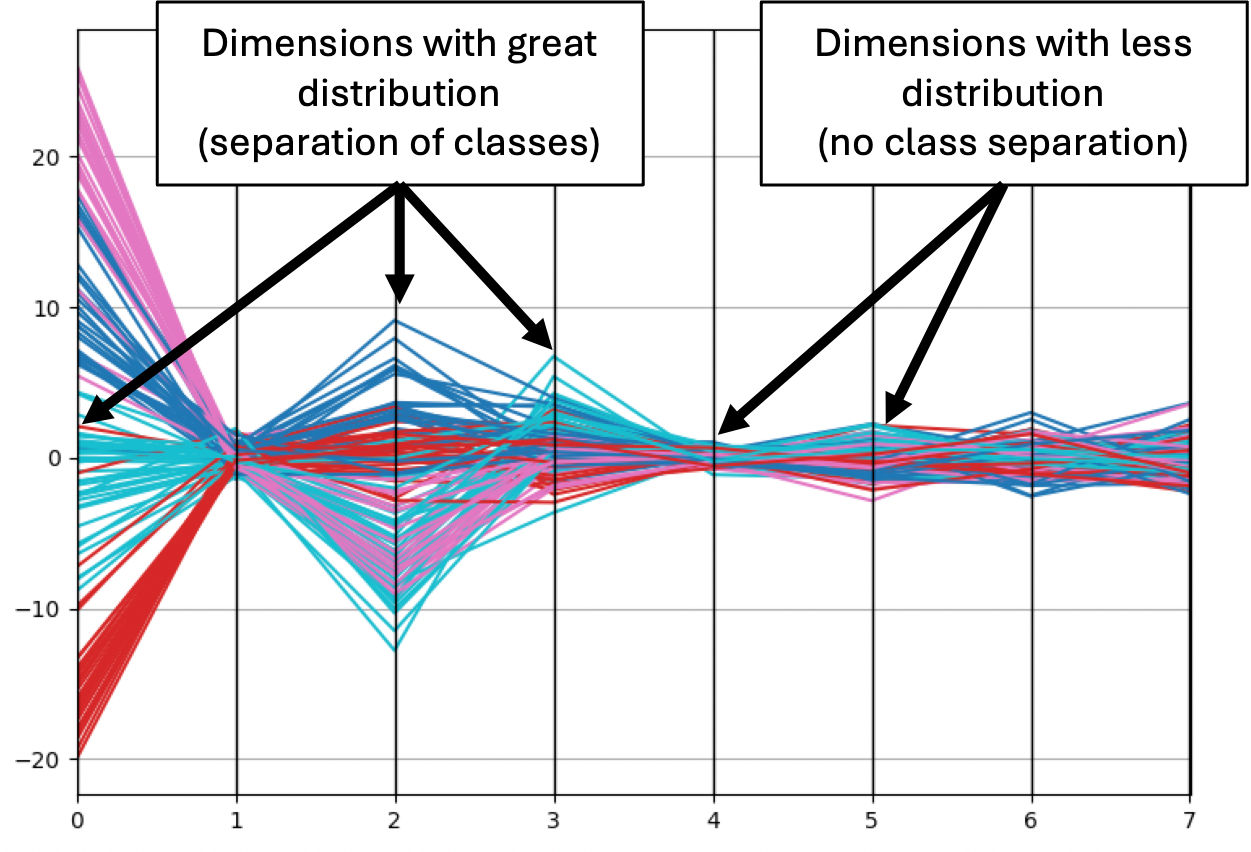}
    \caption{Parallel Coordinates Plot to investigate the relevance of each dimension.}
    \label{fig:parallel}
  \end{subfigure}
  \caption{Potential use cases of latent space visualizations for human-in-the-loop interaction and optimization of the machine learning training.}
  \vspace{-10pt}
  \label{fig:plots}
\end{figure}

Within this work, Hybrid Intelligence forms the core to leverage extended knowledge.
The general idea is to combine human expertise with artificial intelligence to integrate such knowledge directly into the system.
The training of machine learning models usually does not provide any deep insights, acting mostly as a black-box system.
Even though statistical features can be extracted from the training, such as loss or accuracy, there is still a lack of visualizations on how the training data itself is propagated through the model.

As shown in \Cref{fig:plots}, such visualizations can be implemented through two general latent space exploration strategies of the model layer outputs.
Firstly, this can involve the use of high-dimensional visualization techniques such as radar plots or parallel coordinates plots as shown in \Cref{fig:parallel}. 
The advantage of such visualization correlates with the intention of this work to inspect the latent space on the data level.
Due to the plotting of all available dimensions, the user can easily distinguish the importance of each dimension and how the data spreads within one dimension.
As an interactive step, different classes can be dragged apart by moving them on the dimensions axis.
This information may then be fed back into the model to help convergence of the training.

Secondly, dimension reduction methods \cite{reddy2020analysis} like Principle Components Analysis can provide useful insights in the form of scatter plots.
However, reducing dimensions from latent spaces leads to a loss of information which may lead to insufficient interpretation.
Additionally, dimension reduction techniques like t-SNE or UMAP are unable to invert the transformations, even though they are particularly effective in capturing the structure of high-dimensional data for visualization purposes. 
Therefore, the possibilities of those techniques are limited to visual analysis.
As shown in \Cref{fig:scatter}, the scatter plots can still be utilized to organize outliers through drag and drop commands or even move close clusters away from each other.
The distance information can be incorporated through the HITL system for training optimization.

The data-oriented insight offers isolated analysis of data points to investigate if the data or the model itself leads to decreased performance.
Especially for HAR, correlations between sensor data and classes can be derived to enhance the understanding of training procedures.

Connecting within the training loop, such a HITL system enables experts to intervene during the training by visualizing the data and highlighting inefficiencies.
Based on that, advanced adjustment strategies can be implemented through modification of the latent space visualization.
Classic user interface inputs such as marking, selecting, or dragging data points can be utilized to set focal points in the model training to generate a dynamic training concept.
In the sense of this work, such dynamic adjustments can leverage human expertise to especially fine-tune and optimize the training towards efficient resource utilization but are not limited to this use case.

LLM agents such as the popular Chat-GPT or Llama can further support this process by providing smart suggestions during training, assisting the human in the interactive process.
Especially when utilizing multi-modal LLMs, such as VisionLLaMA \cite{chu2024visionllama}, the agent may have an enhanced understanding of the problem since it is able to investigate the problem visually and numerically, especially if visualizations exceed human interpretability and complexity.
The agent can therefore either assist the HITL system or even operate on its own depending on the user's choice.

\subsection{Energy Awareness}

Effective energy tracking is crucial for sustainable machine learning to quantify the expenses of computational resources.
As discussed, tools like Carbontracker \cite{anthony2020carbontracker} or eco2Ai \cite{budennyy2022eco2ai} can be utilized to monitor the energy consumption of models during training. 
However, current issues include the lack of comprehensive energy tracking, as some tools only monitor specific components like the GPU. 
Especially secondary energy consumption, for instance from cooling, is commonly neglected in the tracking even though it contributes to the overall energy investment.
To address this, we propose integrating energy awareness directly into the training loop. 
This involves using sensors such as hall sensors to confirm deviations in energy usage and provide accurate measurements of the full system by gathering the energy consumption directly from the power cord.
By incorporating these energy metrics into the hyperparameter optimization process, we ensure that models are trained with energy efficiency in mind. 
Furthermore, utilizing it in combination with Hybrid Intelligence provides explainability and insights into the energy consumption patterns of different model and training components.

\subsection{Framework Integration}

The final outcome of this work targets to integrate the Hybrid Intelligence and the energy awareness approaches into a robust framework that connects all components seamlessly. 
A user-friendly frontend tool is essential for interacting and monitoring the training process, helping to visualize the training process, track resource usage, and allow for user interaction.

The aforementioned visualizations should be concise and interpretable to allow proper HITL interaction.
Therefore, a generalized and flexible architecture will adapt to the machine learning problem, the model architecture, and the user's goals.
The smart agent can be added as an assistant next to the HITL in order to detect ambiguities and provide suggestions for sustainable adaptations.
Multiple scenarios are possible to either give the human or the smart agent more room for interaction depending on the agent's capabilities.

The energy-awareness approach will not only be added through simple monitoring into the framework, moreover, the energy information can be utilized to optimize the training process by integrating the energy information through penalization of inefficient training states.
Further, resources can be artificially limited or extended during training to force the training process to run more efficiently.
For instance, Hybrid Intelligence can isolate the bad-performing parts of the training data and set priorities to utilize the available resources solely for resolving the current issues.
From that, we can estimate the energy savings and generate forecasts for the total energy investment for training the model.
In the long term, such a framework can be extended beyond pure training adaptation in order to provide a lifecycle-oriented toolkit for estimating energy investments from initial idea to final deployment.

\section{Evaluation}
We envision evaluating our approach through a comprehensive two-stage process with an extended focus on machine learning applications from the HAR field.

First, a numerical analysis of the machine learning problem from data, architecture, and resource investment perspectives.
This involves a detailed examination of data quality impacts, such as sensor placement variability, noise, and labeling errors, on the performance of the machine learning models. 
By identifying these issues from common HAR datasets, we aim to quantify their effects on model accuracy and efficiency with respect to the utilized model architecture.
Further, we will measure and compare the energy consumption across various stages of the machine learning pipeline. This includes data collection, preprocessing, model training, and inference. We will evaluate energy usage with and without the integration of hybrid intelligence interventions to demonstrate potential energy savings and improved efficiency.

Secondly, we investigate the effectiveness of HITL from our Hybrid Intelligence approach.
We will gather feedback from human experts who participate in the HITL process. 
Such feedback will help us assess the effectiveness of real-time adjustments and optimizations made during the model training and deployment phases. 
Additionally, the extension from HITL to LLM agents can help collect user feedback to determine the perceived value and satisfaction with the LLM interventions. 
This evaluation will focus on how smart agents assist non-experts in generating an optimized development environment.

By combining numerical analysis with user feedback, we aim to provide a thorough evaluation, demonstrating the potential of the final framework to enhance both the sustainability and performance of machine learning models concurrently.

\section{Contribution}
The main goal of this work is to connect smart and interactive visualizations with sustainability approaches to reduce computational resource requirements. Our contributions can be divided into two main research directions as discussed before, the Hybrid Intelligence visualizations and the energy awareness.

Building on previous work in latent space exploration as described in \cite{geissler2023latent}, this research initially conducted an asynchronous analysis of model layer outputs. 
The next step is to make this process more interactive by incorporating real-time visualizations and user feedback within the training directly.
After extending this work with an LLM-based agent, future research aims to compare the effectiveness of human versus LLM agents in terms of user experience and overall effectiveness in optimizing the training process.

Towards energy awareness, the impact of hyperparameter settings and different hardware environments on energy consumption has already been investigated through \cite{geissler2024power}. 
Ongoing research is focused on embedding energy consumption metrics directly into the training process. 
One approach being explored is energy-aware hyperparameter optimization using methods like the successive halving algorithm. 
This integration aims to optimize model training not just for performance but also for energy efficiency.

In the field of ubiquitous computing, particularly within HAR, the reliance on time series sensor data from real-world applications presents several challenges. 
Wearable sensing technologies often suffer due to their real-world nature gathered from human activities, such as sensor quality, placement variability, environmental noise, inconsistent experimental procedures, and error-prone data labeling. 
These challenges can significantly degrade machine learning model performance, leading to inefficient resource usage as attempts are made to compensate, often through increased model complexity.
Future research will focus on state-of-the-art datasets from HAR to identify ambiguities within the data itself. 
By adapting models and training processes to be aware of these data issues, we aim to improve overall efficiency. 
The Hybrid Intelligence approach is crucial here, as it allows us to bridge the gap between energy-aware model optimization and HITL systems in the scenario of HAR.

\section{Conclusion}

By leveraging the strengths of human cognition and artificial intelligence, Hybrid Intelligence can significantly improve the sustainability and effectiveness of ubiquitous applications, addressing the various challenges from data, models, and hardware.
Most of the introduced approaches are still in the conceptualization phase and require in-depth investigation through future work.

\begin{acks}
This work is supported by the European Union’s Horizon Europe research and innovation program (HORIZON-CL4-2021-HUMAN-01) through the "SustainML" project (grant agreement No 101070408).
\end{acks}

\newpage

\bibliographystyle{ACM-Reference-Format}
\bibliography{ref}

%%
%% If your work has an appendix, this is the place to put it.

\end{document}